\RequirePackage{fix-cm}
\documentclass[smallextended]{svjour3}       
\smartqed  
\usepackage{graphicx}
\usepackage{amssymb}
\usepackage[latin1]{inputenc}
\usepackage{hyperref}
\usepackage{setspace}
\usepackage{amsmath}
\usepackage{url}
\usepackage{cite}
\usepackage{amstext}
\usepackage[nearskip,caption=false,font=footnotesize]{subfig}

\usepackage{color}
\definecolor{greencolor}{rgb}{0,0.5,0.2}
\definecolor{redcolor}{rgb}{0,0,0}
\definecolor{bluecolor}{rgb}{0,0.,1.}
\definecolor{greycolor}{rgb}{.5,.5,.5}
\def\Red#1{{\color{redcolor} #1}}

\begin{document}

\title{Comparing the \Red{topological properties} of real and artificially generated scientific manuscripts}

\titlerunning{Comparing the writing style of real and artificial papers}        

\author{Diego Raphael Amancio}


\institute{Diego Raphael Amancio \at
Institute of Mathematics and Computer Science \\
University of S\~ao Paulo, P. O. Box 369, Postal Code 13560-970 \\
S\~ao Carlos, S\~ao Paulo, Brazil\\
}

\date{Received: date / Accepted: date}

\maketitle

\begin{abstract}
Recent years have witnessed the increase of competition in science. While promoting the quality of research in many cases, an intense competition among scientists can also trigger unethical scientific behaviors. To increase the total number of published papers, some authors even resort to software tools that are able to produce grammatical, but meaningless scientific manuscripts. Because automatically generated papers can be misunderstood as real papers, it becomes of paramount importance to develop means to identify these scientific frauds. In this paper, I devise a methodology to distinguish real manuscripts from those generated with SCIGen, an automatic paper generator. Upon modeling texts as complex networks (CN), it was possible to discriminate real from fake papers with at least 89\% of accuracy. A systematic analysis of features relevance revealed that the accessibility and betweenness were useful in particular cases, even though the relevance depended upon the dataset. The successful application of the methods described here show, as a proof of principle, that network features can be used to identify scientific gibberish papers. In addition, the CN-based approach can be combined in a straightforward fashion with traditional statistical language processing methods to improve the performance in identifying artificially generated papers.
\end{abstract}

\section{Introduction}

The dissemination of knowledge and the advancement of science strongly depend upon the precise interpretation of the content conveyed in scientific manuscripts. Therefore, the ideas conveyed by high-quality scientific papers should be carefully detailed so that they can be tried and possibly improved. Although many qualitative aspects have been proposed to identify outstanding manuscripts and their respective authors, many quantitative aspects still prevail when the quantification of academic merit is at stake. For example, in recent years, the total number of articles, the number of citations motivated by articles or researchers' h-index has been widely used for the purpose of merit evaluation~\cite{evaluacao}. Clearly, there is a correlation between quantitative and qualitative factors~\cite{acorrelacao}. Nevertheless, the drawbacks related to the exclusive use of quantitative factors are well known. For example, recent publications cannot be evaluated via citation counts. Likewise, very young researchers also cannot be assessed according to the number of citations that their articles motivate, since there is an expected delay between paper publication date and wide scientific recognition~\cite{ex1,ex2,ex3,ex4,ex5}.

While there are several disadvantages associated with quantitative indices, they unmistakably provide a minimal degree of objectivity required for any scientific merit assessment. Aware of the prevalence of quantitative indexes in scientific merit judgments, some scholars tend to shape their research only to increase their citation counts and other quantitative indexes~\cite{selfie1}. The pressure imposed by scientific competition, translated by the maxim ``publish or perish'', literally urge a few scholars not to follow good scientific practices. In order to artificially boost productivity and impact indexes, some authors split the results arising from a single discovery in two or more papers.
For these reasons, several scientific low-quality papers with a very weak impact on science have been produced. Other recurrent unethical conducts include the excessive use of self-citations~\cite{selfie1,selfie2} and plagiarisms~\cite{plagio1,plagio2,plagio3}.
More recently, even texts automatically generated have been submitted and surprisingly deemed suitable for publication in several scientific conferences~\cite{labbe2010}. Currently, one of the most popular software for generating fake papers is the SCIGen~\cite{sciref}, an algorithm able to produce gibberish papers that resemble real manuscripts. To produce such meaningless texts, SCIGen uses a complex grammar that is able of generating texts containing all the features expected in a standard scientific manuscript. To complement the grammar, even figures and tables are generated. An incremental modification of the original algorithm has implemented the possibility of self-citations, which has allowed a significant increase in authors' h-index~\cite{hindman}. Because fake papers as those generated by SCIGen can eventually bewilder even a human referee, it becomes of paramount relevance the identification of particular features able to discriminate real from meaningless texts. In this context, I focus on one approach to identify distinct styles in texts that has proven particularly effective to detect SCIGen texts. \Red{More specifically, using a representation of texts as complex networks~\cite{newman}, I show that it is possible to discriminate real and fake manuscripts with significant accuracy if one analyzes the structural organization of the manuscripts. Even though the accuracy of the proposed technique does not outperforms other traditional methods based on the analysis of textual content, it is useful to show the structural patterns of text organization is affected when fake information is conveyed.
}

\Red{This manuscript is organized as follows. In Section \ref{s2}, I present related approaches aiming at the identification of fake scientific manuscripts. A very short introduction to the application of complex networks for text analysis is presented is Section \ref{s3}. The methods employed for the representation, characterization and classification of text networks are presented in Section \ref{smethods}. The results obtained with both univariate and multivariate analysis of network measurements are presented in Section \ref{sresults}. Finally, the conclusion drawn from the results and the perspectives for further studies are commented in Section \ref{sconclusion}.
}

\section{Related works} \label{s2}

\Red{
Several methods have been devised to identify the authenticity of scientific manuscripts. Such methods can be classified according to the type of information that is employed as features of classifiers. Usually the list of references plays a important role in the task. For example, it has been shown that when many cited references cannot be found online, then there is a high probability that the paper under analysis is fake~\cite{xiang}.
}

\Red{
Many heuristics rely on textual content to infer the authenticity of documents~\cite{ginsparg,noorden}. The study developed in~\cite{lavoie} proposes some useful rules. One of the main rules tests whether keywords in the title and abstract occurs frequently in the body of the paper. If such pattern does not occur, then the document is considered as fake.
Another interesting observation highlights that real scientific papers usually mention keywords from the titles of cited papers. Techniques based on the semantic content of texts also employ traditional similarity measurements~~\cite{statnlp}. An important contribution to the problem of identifying gibberish publications with similiraty measurements was introduced in~\cite{labbe2013}. In their study, the authors devised a pairwise similarity measurement that compares two pieces of texts by counting differences in word frequencies. This approach was  useful to identify several cases of duplicate and fake publications. An extension of this work was proposed in~\cite{estrutural}, where not only single word occurrences are considered, but also multi-word phrases. Other interesting approaches
relying on textual content include the techniques based on the compressibility rate of texts~\cite{li,compressao}. In~\cite{compressao}, the authors show that artificially generated papers display values of compressibility rate that are not compatible with the rates observed in real manuscripts.
}

\Red{
Differently from approaches mentioned in this section, the method I proposed in this paper does not consider the \emph{semantic} similarity of texts. Actually, the approach proposed here focus on the analysis of connectivity patterns that are able to capture subtleties of styles in texts from distinct sources. Because the proposed approach is complementary to other traditional techniques, it could potentially be useful to improve the reliability of the classification.
}

\section{Complex networks and text analysis} \label{s3}

\Red{
Complex networks have been employed to model a myriad or real complex systems~\cite{newman}. Of particular interest to the aims of this study are the applications in automatic summarization~\cite{extractive,lantiq}, machine translation~\cite{trad1,trad2}, complexity analysis~\cite{complexity1,complexity2,complexity3,complexity4,complexity5} and authorship recognition~\cite{njp,plosrecent}. In all these tasks, the networks obtained from the so called word adjacency model (see Section \ref{wnmodel}) allowed a precise characterization of texts with regard to specific textual features. Networked models even allowed the characterization of unknown manuscripts~\cite{probing} and many other linguistic aspects~\cite{lang2,lang3,lang4,lang5,lang6}. Interestingly, the particular features of each language could also be classified in terms of the topological structure of complex networks~\cite{doliu1,domehler,doliu2}. A more detailed survey on the application of network methods in text analysis can be found in~\cite{lang1}. Differently from traditional approaches, the method proposed here focus on the structure and organization of texts, rather than on the textual semantic content. The proposed method also differs from the other techniques mentioned here because it is modified in order to
to minimize the influence of the vocabulary size on the topological analysis of scientific articles (see Section \ref{sresults}).
%
}

\section{Methodology} \label{smethods}

\begin{figure}[!h]
\centering
\includegraphics[width=0.9\linewidth]{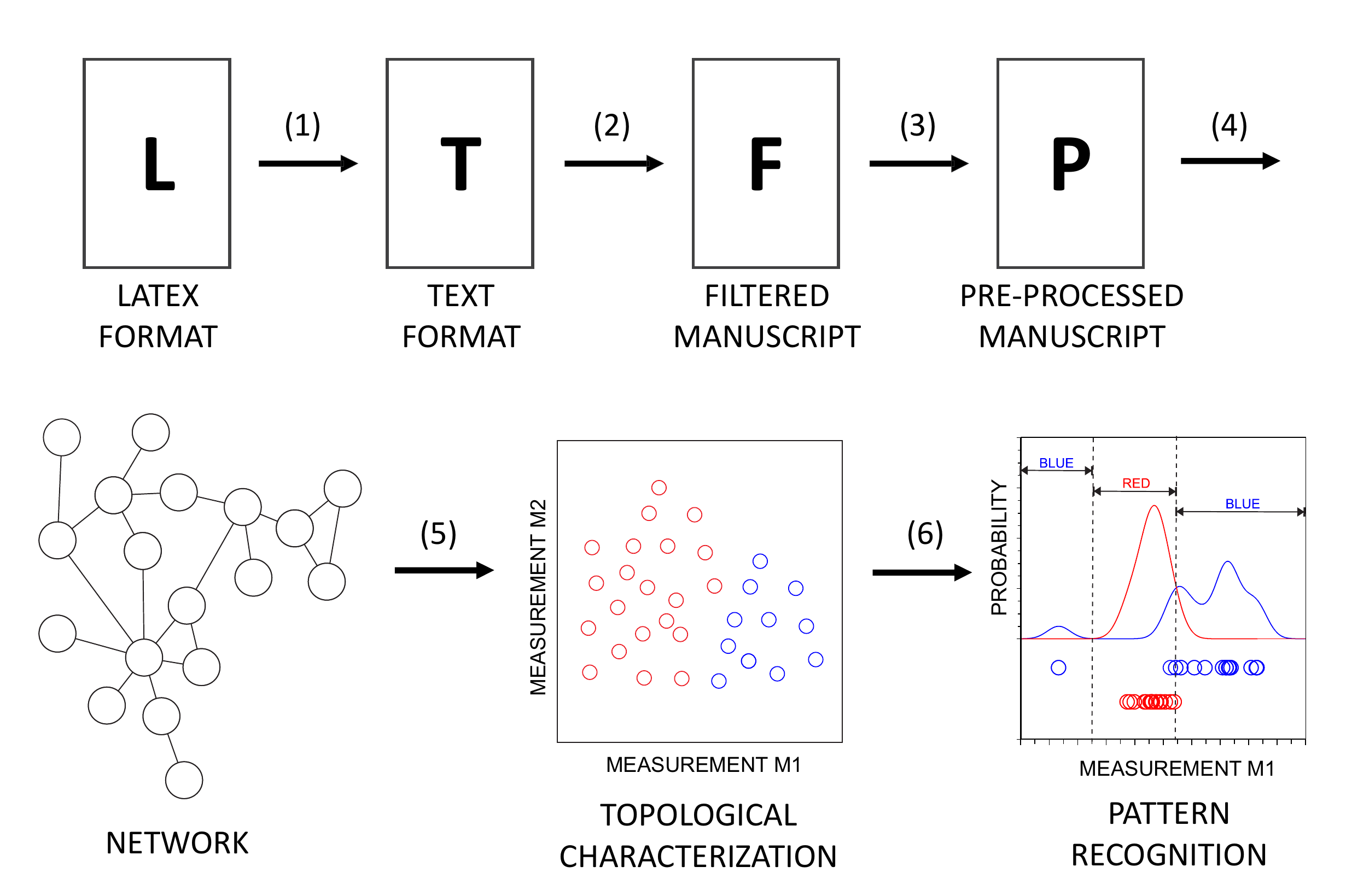}
\caption{Sequence of methods employed to distinguish gibberish from real scientific manuscripts. The actions taken in each step are: (1) \LaTeX\ tags and mathematical terms are stripped out; (2) the manuscript is manually checked in order to verify if its content includes only textual information; (3) lemmatization and removal of \emph{stopwords}; (4) mapping of a text into a network; (5) extraction of complex network measurements; (6) discrimination of distinct classes (real or fake) via machine learning. }
\label{fig.metodos}
\end{figure}

The methodology employed to compare real and fake manuscripts is illustrated in Figure \ref{fig.metodos}. Firstly, the text of the scientific article are obtained by automatically removing the \LaTeX\ tags from the original texts. Because some undesirable tokens can still remain in the text, the output is checked manually.
Following previous studies, the style of each text is quantified via topological characterization of complex networks (graphs)~\cite{probing,njp,plosrecent,sigman,syntactic}. For this reason, the texts are modeled as complex networks. Then, several connectivity measurements are extracted from the networks. In the next step, the measurements are employed as features to discriminate real and fake manuscripts. The discrimination is accomplished with pattern recognition methods. The main steps shown in Figure \ref{fig.metodos} are described below.

\subsection{Modeling texts as complex networks} \label{wnmodel}

A network can be defined as a set of nodes connected by edges. To represent a network, consider that $A = \{a_{ij}\}$ is the matrix representing the network structure. In texts, each distinct word is a node and edges are established between adjacent words. Therefore, if words $i$ and $j$ appear adjacent in the text, the element $a_{ij}$ is set ($a_{ij}=1$). Otherwise, $a_{ij}=0$. The total number of links, i.e. the node degree, is defined as $k(i) = \sum_j a_{ij}$.
In several style-based applications, some pre-processing steps are usually applied before the connection of adjacent words~\cite{lemmatisation}. The pre-processing algorithm encompasses a two-fold mechanism: (a) the identification and removal of \emph{stopwords}; and (b) the lemmatization. In (a), words conveying low semantic content (e.g. ``and'', ``of'', ``a'', ``an'') are removed. Since these words are simply used to connect content words, they can be straightforwardly replaced by edges. In (b), each remaining word is mapped to its canonical form~\cite{statnlp}. As a consequence, conjugated verbs are mapped to their infinitive forms. Likewise, nouns are mapped to their singular forms. In order to obtain word lemmas~\cite{statnlp}, it is imperative to know in advance the part-of-speech of words. In this study, each word was labeled with its part-of-speech using a maximum entropy model~\cite{maxent}. To illustrate the modeling of a text as a network, the pre-processing steps (a) and (b) are applied to a short extract from the book ``Adventures of Sally'', by P.G. Wodehouse:
\begin{description}

  \item[\ \ \ \ ] {\bf Original text:} ``If Sally had been constantly in Bruce Carmyle's thoughts since they had parted on the Paris express, Mr. Carmyle had been very little in Sally's--so little, indeed, that she had had to search her memory for a moment before she identified him''.

\end{description}

\begin{description}
  \item[(a)] {\bf Removal of stopwords and punctuation marks:} Sally constantly  Bruce Carmyle thoughts parted  Paris express Carmyle little Sally little search  memory  moment before identified

  \item[(b)] {\bf Lemmatization:} Sally constant Bruce Carmyle think part Paris express Carmyle little Sally little search memory moment before identify

\end{description}

\begin{figure}[!h]
\centering
\includegraphics[width=0.75\linewidth]{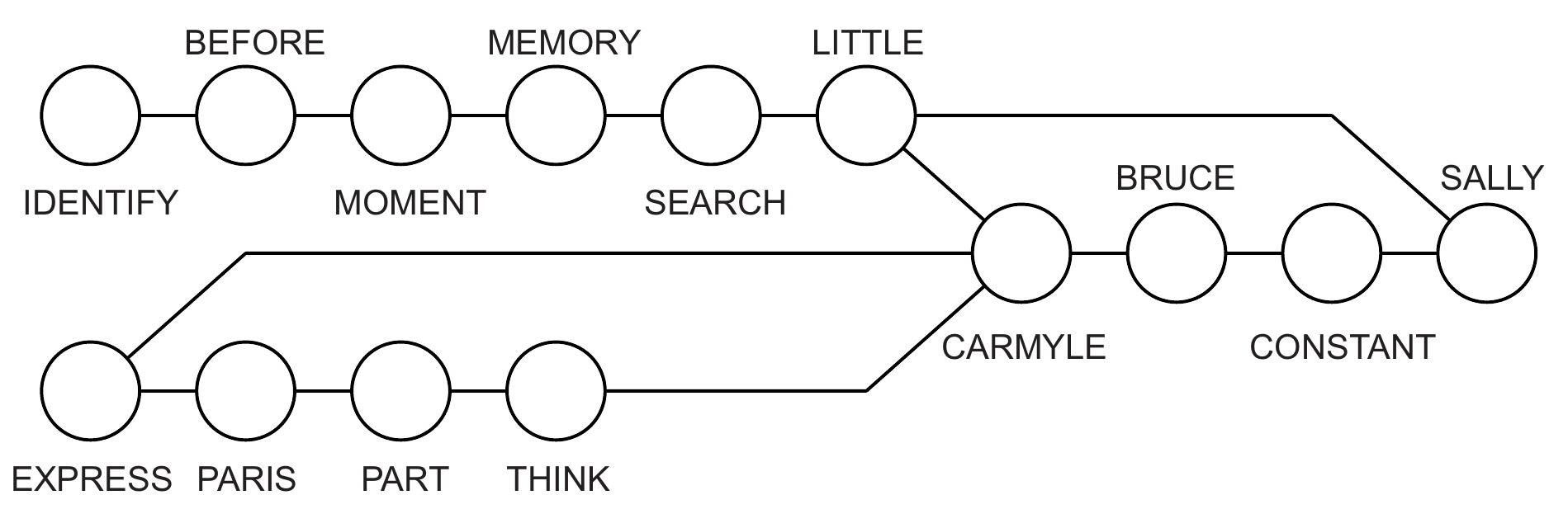}
\caption{Example of network obtained from the text: ``If Sally had been constantly in Bruce Carmyle's thoughts since they had parted on the Paris express, Mr. Carmyle had been very little in Sally's--so little, indeed, that she had had to search her memory for a moment before she identified him''.}
\label{fig.net}
\end{figure}

The network obtained from the pre-processed text is shown in Figure \ref{fig.net}. Note that, after step (b), each distinct word becomes a distinct node and edges are established between adjacent words.

\subsection{Topological measurements of complex networks}
\label{netm}

The topological characterization of complex networks can be performed by computing topological measurements. Currently, there exists several topological measurements~\cite{newman}. In this study, the measurements usually employed for textual analysis were chosen to characterize the topological attributes of text networks. A swift description of each measurement is provided below.
\begin{itemize}

  \item {\bf Average node degree}: this local measurement, quantifies the average connectivity of the neighbors:
      \begin{equation}
        k_n(i) =   \sum_j a_{ij} k(i) ~ \big{/} ~ \sum_j a_{ij}.
      \end{equation}

  \item {\bf Clustering coefficient}: the clustering coefficient (C) is a quasi-local measurement that quantifies the density of links between neighbors.
      Mathematically, the clustering coefficient is defined as $C = 3 n_a / n_b$, where
      \begin{equation}
        n_a = \sum_{k>j>i} a_{ij} a_{ik} a_{jk},
      \end{equation}
      \begin{equation}
        n_b = \sum_{k>j>i} a_{ij} a_{ik} + a_{ji} a_{jk} + a_{ki} a_{kj}.
      \end{equation}
      In textual applications, the clustering coefficient of specific words tends to quantify the number of distinct contexts in which the word appears~\cite{njp}.

  \item {\bf Accessibility}: the accessibility (or diversity) ($\alpha$) is a extension of the degree that is based on both  topology and dynamics of networks~\cite{extractive}. This centrality index is relevant to identify topological high-degree nodes that effectively access only a few neighbors~\cite{border}.
      To define this measurement, consider the following definition. Let $p_{ij}^{(h)}$ be the likelihood of a random walker to go from node $i$ to node $j$ in $h$ steps. The accessibility is computed as the irregularity of the distribution of $p_{ij}^{(h)}$:
      \begin{equation}
        \alpha^{(h)}(i) = \exp \Big{(} - \sum p_{ij}^{(h)} \ln  p_{ij}^{(h)} \Big{)}.
      \end{equation}
      This measurement has been employed to detect the border of complex networks~\cite{border}. In textual applications, the accessibility has been useful to identify core concepts, allowing thus the construction of informative automatic summarizers~\cite{extractive}.

  \item {\bf Average shortest path length}: the shortest path length ($l$) quantifies the typical distance between two nodes of the network. This measurement was employed because it has been useful in textual applications~\cite{njp,identification}. In word adjacency networks, this measurement has proven relevant to identify keywords, even if they are not very frequent~\cite{njp}.

  \item {\bf Betweenness}: the betweenness ($B$) is a centrality measurement. This means that the highest values of betweenness are assigned to the most relevant concepts in word adjacency networks. This measurement quantifies how easily a node can be accessed, provided that walks are performed via shortest paths. Let $g_{ij}^{(m)}$ be the number of shortest paths between nodes $i$ and $j$ passing through node $m$. If $g_{ij}$ is the total number of shortest paths between $i$ and $j$ (passing through any intermediary node), then the betweenness is defined as
      \begin{equation}
        B(i) = \sum_i \sum_j \frac{g_{ij}^{(m)}}{g_{ij}}.
      \end{equation}
      In textual networks, the betweenness quantifies the number of distinct contexts of a given word~\cite{njp}. Unlike the clustering coefficient, this measurement uses the global network connectivity to infer the number of contexts~\cite{njp}.

  \item {\bf Assortativity}: the  assortativity ($r$) quantifies degree-degree correlations~\cite{mixing}. In other words, it measures the tendency of nodes with similar degree to be connected. Mathematically, it can be defined as
      \begin{equation}
        r = \frac{M^{-1} \sum_{j>i} k(i) k(j) - \Big{[} M^{-1} \sum_{j>i} a_{ij} (k(i) + k(j))/2 \Big{]}^2}
        {M^{-1} \sum_{j>i} a_{ij}( k^2(i) + k^2(j) )/2 - \Big{[} M^{-1} \sum_{j>i} a_{ij} (k(i) + k(j) )/2) \Big{]}^2 }.
      \end{equation}
      Networks whose assortativity take positive values are referred to as assortative networks. On the other hand, if there is a negative correlation between the degree of linked nodes, then the network is disassortative. In word adjacency networks, a disassortative behavior arises even when stopwords are removed from the analysis~\cite{finding}.

\end{itemize}

\subsection{Pattern recognition methods}

In a supervised classification task, the objective is to automatically distinguish objects (or instances) according to their classes. The characterization of each object is made with object attributes (or features). In this study, one desires to distinguish between two class: (i) the ``real'' class, which include real scientific papers; (ii) and the ``fake'' class, which encompasses
the papers automatically generated by the SCIGen algorithm. As features, I chose the network measurements described in Section \ref{netm}.
The following pattern recognition methods were employed in this study:

\begin{itemize}

  \item {\bf Naive Bayes (NBY):} the naive bayes classifier uses the Bayesian optimal decision rule to classify an object. The class $c'$ is chosen if the condition
\begin{equation}
    P(c'|m) > P(c_k|m),
\end{equation}
holds for each $c_k \neq c'$, where $P(c_k|m)$ is the likelihood of class $c_k$ to appear in the context represented by the set of network measurements $m$. In most cases, the exact behavior of $P(c_k|m)$ is unknown. To overcome this issue, the Bayes' theorem can be used:
\begin{align} \label{bayestheo}
    c' & = \arg \max_{c_k} P(c_k|m)  = \arg \max_{c_k} \frac{P(m|c_k)}{P(m)} P(c_k) \nonumber \\
       & = \arg \max_{c_k} P(m|c_k)  P(c_k) = \arg \max_{c_k} \big{[} \log P(m|c_k) + \log P(c_k) \big{]}.
\end{align}
Assuming in eq. (\ref{bayestheo}) attribute independence and considering that the topological context is given by a set of network measurements $m=\{m_1,m_2\ldots\}$, then $P(m|c_k)$ can be written as
\begin{equation}
    P(c_k|m) = P(\{m_i | m_i \in m\} | c_k ) = \prod_{m_i \in m} P(m_i|c_k).
\end{equation}
Therefore, the accurate class $c'$ associated to the unknown instance is
\begin{equation} \label{decisao}
    c' = \arg \max_{c_k} \big{[} \log P(c_k) + \sum_{m_i \in m} \log P(m_i|c_k) \big{]}.
\end{equation}

To illustrate the decision process, consider Figure \ref{fig.naive}. The position of each circle in the x-axis represents the value obtained for a given measurement. Distinct colors represent different classes ($c_1$=``blue'' and $c_2$=``red''). Considering that the frequency of occurrence in the dataset of each class is the same, the term  $\log P(c_k)$ can be disregarded in eq. \ref{decisao}. The remaining term, the likelihood  $P(m_i|c_k)$, can be estimated via the Parzen window method~\cite{parzen}. Therefore, the decision boundaries are established according to
\begin{equation}
    c' = \arg \max_{c_k}  P(m|c_k).
\end{equation}

\begin{figure}[!h]
\centering
\includegraphics[width=0.32\linewidth]{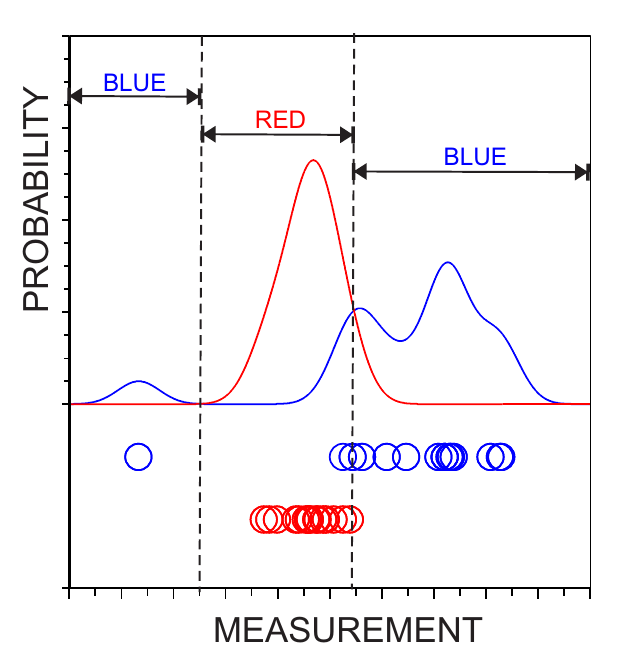}
\caption{Example of classification between two classes (red and blue) using the Naive Bayes algorithm. The probability distribution of each class is used to create decision boundaries.}
\label{fig.naive}
\end{figure}

\item {\bf Nearest neighbors (KNN):} in this algorithm, the classification of an unknown instance is performed with a voting process which considers the k nearest neighbors. If most of the k nearest neighbors belong to the class $c_k$, then $c_k$ is associated to the unknown instance. In Figure \ref{fig.knn}, the innermost dashed circle represents the set of instances used for the voting process when $k=5$. In this case, the class associated to the unknown instance represented by a question mark (?) is the red class. If the $k=12$ nearest neighbors are chosen for the voting process, the most frequent class becomes the blue class. Finally, if $k=19$, the class associated to the unknown instance is the blue class. In this paper, the value $k=1$ was used since this value usually provides highest accuracy rates~\cite{sistematica}.

\begin{figure}[!h]
\centering
\includegraphics[width=0.30\linewidth]{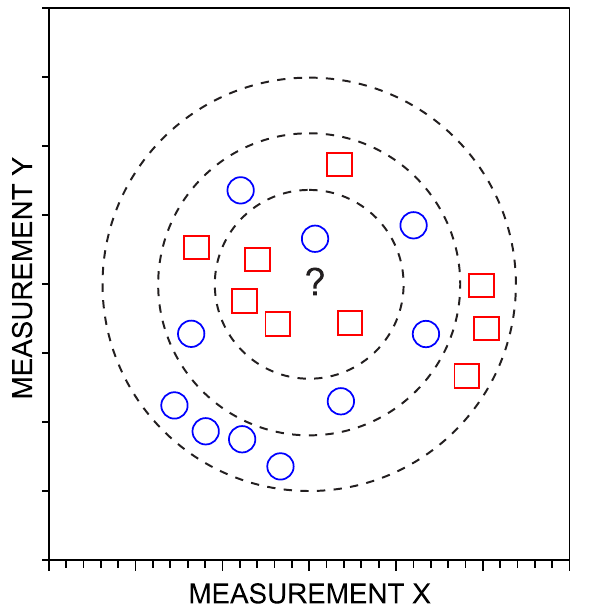}
\caption{Classification of the unknown instance (central question mark) using the kNN algorithm. If the innermost dashed circle is used ($k=5$), then the class associated to the unknown instance is the reddish one.}
\label{fig.knn}
\end{figure}

\item {\bf Decision trees (C45):} this method uses a tree as a data structure to represent the emergent patterns of the dataset~\cite{cormen}. More specifically, in a decision tree, each node represents an attribute and edges correspond to tests performed on attributes (see Figure \ref{aarv}). The decision process starts at the root (i.e. the node with no parents). When a leaf node is reached, the class associated to that node is selected. The generation of a decision tree requires the definition of a measure that is able to identify the most informative attribute at each step of the algorithm. More specifically, in this paper, I used the Kullback-Leibler divergence~\cite{duda}. Mathematically, the Kullback-Leibler divergence $\Omega( \mathcal{S}_{tr}, m_i )$ of the attribute $m_i$ computed in the training dataset $\mathcal{S}_{tr}$ is
    \begin{equation}
        \Omega( \mathcal{S}_{tr}, m_i ) = \mathcal{H}(\mathcal{S}_{tr}) - \mathcal{H}(\mathcal{S}_{tr}|m_i),
    \end{equation}
    where $\mathcal{H}(\mathcal{S}_{tr})$ is the entropy computed in the the training dataset $\mathcal{S}_{tr}$ and $\mathcal{H}(\mathcal{S}_{tr}|m_i)$ is the entropy of the dataset when the value of $m_i$ is specified. Particularly, $\mathcal{H}(\mathcal{S}_{tr}|m_i)$ can be computed from $\mathcal{S}_{tr}$ as
     \begin{equation}
                 \mathcal{H}(\mathcal{S}_{tr}|m_i)  =   \sum_{v \in V(m_i)}   \frac{ | \beta_{(tr)} \in \mathcal{S}_{tr} |  \beta_{(tr)}^{(k)} = v | }{ | \mathcal{S}_{tr} | } \cdot
                  \mathcal{H}( \{  \beta_{(tr)} \in \mathcal{S}_{tr} |  \beta_{(tr)}^{(k)} = v  \},
     \end{equation}
     where $V(m_i)$ is the set of all values taken by the attribute $m_i$ in the training dataset.

     \begin{figure}[!h]
     \centering
        \includegraphics[width=0.42\linewidth]{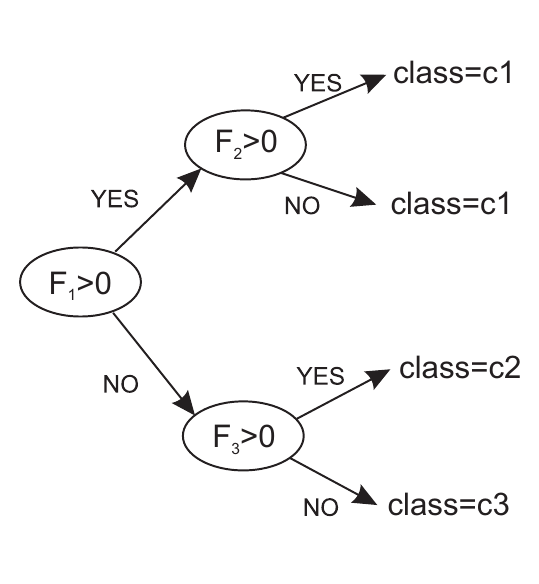}
        \caption{Example of a decision tree. To decide the class of a unknown instance, consider that attributes are $F_1 = -0.10$, $F_2 = 0.11$ and $F_3 = 0.38$. The decision process starts at the leftmost node, the root. The first test leads to the edge ``NO'' and the second edge leads to the edge labeled as ``YES''. Therefore, the class associated to the unknown instance is the class $c_2$. }
        \label{aarv}
     \end{figure}

\end{itemize}

\subsection{Quantifying feature relevance} \label{qfr}

The method employed for quantifying feature relevance assigns high values of relevance for a given attribute if its use usually yields high quality classifiers. More specifically, this method counts the frequency of appearance of each feature among the best classifiers, when one analyzes all possible combination of features. Let $F$ be a set comprising $n_f$ features. Using $F$, it is possible to generate $n_c = 2^{n_f}$ distinct combinations of features. To quantify the relevance of each feature, the $n_c$ combinations are sorted in decreasing order according to the accuracy rate provided by each combination. Suppose that $\xi_{ij}$ represents the ordered set of combinations, where
\begin{equation}\label{eq.mavg}
    \xi_{ij} = \left\{
    \begin{array}{ll}
        1 & \textrm{ if the $i$-th best combination employed the $j$-th feature}, \\
        0 & \textrm{ otherwise. } \\
    \end{array}
    \right.
\end{equation}
Then $\xi_{ij}$ can be used to verify if a feature $j$ tends to appear among the best classifiers. This can be done by defining the function $f(x)$ as
\begin{equation}
    f(x) = \sum_{i=1}^{x} \xi_{ij}, \ \ \{x \in \mathbb{N}^* | x \leq n_c \}.
\end{equation}
Note that $f(x)$ increases quickly whenever $j$ is frequent among the best combinations of features. Conversely, if a given feature $j$ is more frequent among the worst classifiers, $f(x)$ increases significantly only for high values of the domain. Therefore, the prominence $\rho(j)$ of feature $j$ can be computed as the {area underneath the curve} $f(x)$:
\begin{equation} \label{relevvance}
    \rho(f) = \int_{1}^{n_c} f(x) \mathrm{d}x = \sum_{i=1}^{n_c} \sum_{k=1}^{i-1} \xi_{kj} + \frac{1}{2} \sum_{i=1}^{n_c} \xi_{ij}.
\end{equation}

\section{Results} \label{sresults}

In this study, the style of real and fake manuscripts were compared. As fake manuscripts, I considered the texts generated by the SCIGen algorithm, which produces scientific manuscripts using a proper grammar. The style of the SCIGen papers were compared with the style of real manuscripts recovered from the following sources: (a) the Pattern Recognition Letters journal (PRL)~\cite{prlj}; (b) the arXiv repository comprising Computer Science papers (arXiv/cs)~\cite{arxivj}; and (c) the Journal of Informetrics (JI)~\cite{jinfor}. Four hundred manuscripts were used in the experiments. Note that most of the measurements presented in Section \ref{netm} are local measurements. Therefore, each node is associated to a specific value. To characterize each manuscript, I used the global distribution of measurements for all the words in the manuscript. Here, the goal is to obtain quantities characterizing relevant factors of the distributions to be used as global measurements. Using the same strategy of previous studies~\cite{identification,probing}, the average $\langle X \rangle$ and the deviation $\Delta X$ of each local measurement $X$ was extracted. Therefore, the features employed to characterize the style of the manuscripts were
\begin{equation}
    \langle \alpha^{(h=2)} \rangle,\ \Delta \alpha^{(h=2)},\ \langle \alpha^{(h=3)} \rangle,\ \Delta \alpha^{(h=3)},\ \langle k_n \rangle,\ \Delta k_n,\ \langle B \rangle,\ \Delta B,\ \langle C \rangle,\
    \Delta C,\ r,\ \langle l \rangle,\ \textrm{and }\Delta l. \nonumber
\end{equation}
To minimize the correlation of the above measurements with the frequency of words, the following normalization was applied. Let $\tilde{X}$ be the value of a given measurement obtained in a text and $\langle X^{(\textrm{R})} \rangle$ the average value of the same measurement obtained in $20$ randomized versions of the text. Then, the normalized measurement is computed as
\begin{equation}
    X = \frac{\tilde{X}}{\langle X^{(\textrm{R})} \rangle}.
\end{equation}
After characterizing the topological structure of the manuscripts, the hypothesis that real and fake manuscripts yields distinct network properties was probed. A twofold method was employed to accomplish the identification of fake papers: a univariate and a multivariate approach.

\subsection{Univariate analysis}

In this approach, the discriminability of real and fake papers was analyzed by considering just a single measurement or each classifier generated. The accuracy rate obtained in each dataset is shown in Table \ref{univ.result}. The discrimination of PRL and SCIGen papers was accomplished with an accuracy of 79\% in the best scenario, when the average neighbor degree $\langle k_n \rangle$ and the average accessibility $\langle \alpha^{h=2} \rangle$ was employed along with the tree (C4.5) algorithm. The accurate discrimination between arXiv/cs and SCIGen papers could be performed in 88\% of the cases. This accuracy rate was obtained with the standard deviation of the average neighbor degree $\Delta k_n$. The highest accuracy rate occurred when distinguishing JI and SCIGen papers. In this case, 91\% of the papers could be successfully discriminated. Taken together, these results confirm that the networked representation of texts is useful to distinguish real manuscripts from those automatically generated from SCIGen. Especially, it is possible to note that SCIGen texts are more similar to the PRL manuscripts, which can be explained by both content and structural similarities, because both datasets comprise letters about computer science issues. While the arXiv/cs also comprises Computer Science papers, the format allowed by this repository is much more generic than the structural format generated by the SCIGen algorithm. Hence, as expected, a larger discriminability was found when SCIGen and arXiv/cs were compared. When comparing JI and SCIGen, an even larger distinguishability was obtained probably because both structural and semantical contents are distinct.
%
%
\begin{table}
    \centering
    \caption{\label{univ.result}Accuracy rate (\%) obtained for each measurement in the univariate approach. The best discriminability was found when comparing JI and SCIGen papers.}
    \begin{tabular}{|l|c|c|c|c|c|c|c|c|c|}
      \hline
       & \multicolumn{3}{|c|}{\bf PRL} & \multicolumn{3}{|c|}{\bf arXiv.org/cs} & \multicolumn{3}{|c|}{\bf JI} \\
       \cline{2-10}
       & KNN & NBY & C45 & KNN & NBY & C45 & KNN & NBY & C45 \\
      \hline
      Accessibility $\langle \alpha^{(h=2)} \rangle$ & 72 & 78 & 78 & 74 & 79 & 78 & 74 & 80 & 81 \\
      Accessibility $\Delta \alpha^{(h=2)}$          & 44 & 62 & 48 & 56 & 64 & 55 & 49 & 60 & 49 \\
      Accessibility $\langle \alpha^{(h=3)} \rangle$ & 72 & 77 & 79 & 66 & 73 & 73 & 72 & 76 & 75 \\
      Accessibility $\Delta \alpha^{(h=3)}$          & 44 & 61 & 48 & 83 & 88 & 87 & 86 & 91 & 91 \\
      Avg. N. Degree $\langle k_n \rangle$           & 72 & 77 & 79 & 66 & 63 & 57 & 70 & 78 & 75 \\
      Avg. N. Degree $\Delta k_n $                   & 45 & 62 & 48 & 83 & 86 & 88 & 85 & 90 & 90 \\
      Betweenness $\langle B \rangle$                & 66 & 77 & 78 & 68 & 77 & 77 & 69 & 77 & 74 \\
      Betweenness $\Delta B$                         & 61 & 71 & 66 & 47 & 64 & 64 & 61 & 62 & 58 \\
      Clustering $\langle C \rangle$                 & 50 & 46 & 49 & 57 & 50 & 50 & 50 & 64 & 53 \\
      Clustering $\Delta C $                         & 58 & 58 & 54 & 50 & 55 & 55 & 54 & 68 & 63 \\
      Assortativity $r$                              & 59 & 76 & 74 & 71 & 71 & 72 & 73 & 79 & 78 \\
      Shortest paths $\langle l \rangle$             & 63 & 71 & 65 & 62 & 73 & 71 & 66 & 75 & 69 \\
      Shortest paths $\Delta l$                      & 56 & 58 & 59 & 48 & 64 & 60 & 67 & 68 & 68 \\
      \hline
    \end{tabular}
\end{table}

The individual performance of the attributes employed in the univariate analysis can be summarized as follows:
\begin{itemize}

  \item {\bf Accessibility}: the average accessibility $\langle \alpha^{(h=2)} \rangle$ presented an average discriminative ability. The average accessibility at the third level was particularly useful in the PRL dataset, since the highest accuracy was found when the $\langle \alpha^{(h=3)} \rangle$ was employed with the C4.5 method.   The deviation  $\Delta \alpha^{(h=3)}$ proved specially relevant to identify real papers in the arXiv/cs and JI datasets.

  \item {\bf Neighbors degree}: an excellent performance was found for the $\Delta k_n $ in the arXiv/cs and JI datasets. Conversely, the average $\langle k_n \rangle$ performed well mainly in the PRL dataset.

  \item {\bf Betweenness}: the average $\langle B \rangle$ turned out to be more relevant than the deviation $\Delta B$. Nevertheless, the use of the betweenness as a feature yielded relatively low accuracy rates.

  \item {\bf Clustering coefficient}: this measurement yielded low accuracy rates in all three datasets. This means that the fraction of links between neighbors is not relevant for this task. The most relevant links, therefore, are those connecting neighbors and further hierarchies.

  \item {\bf Assortativity}: in most cases, this measurement presented an average performance.

  \item {\bf Shortest paths}: the average $\langle l \rangle$ was found to be more informative than the deviation $\Delta l$. The best performance achieved with the shortest path length, however, was only 71\%.

\end{itemize}

Even though the univariate analysis is able to identify which attributes are more useful to discriminate specific classes, this analysis does not take into consideration the inter-relationship between different attributes. Because the interaction of attributes may improve the quality of the classifiers, in the next section, I approach the classification task as a multivariate problem.

\subsection{Multivariate analysis}

\begin{figure}[!h]
\centering
\includegraphics[width=0.85\linewidth]{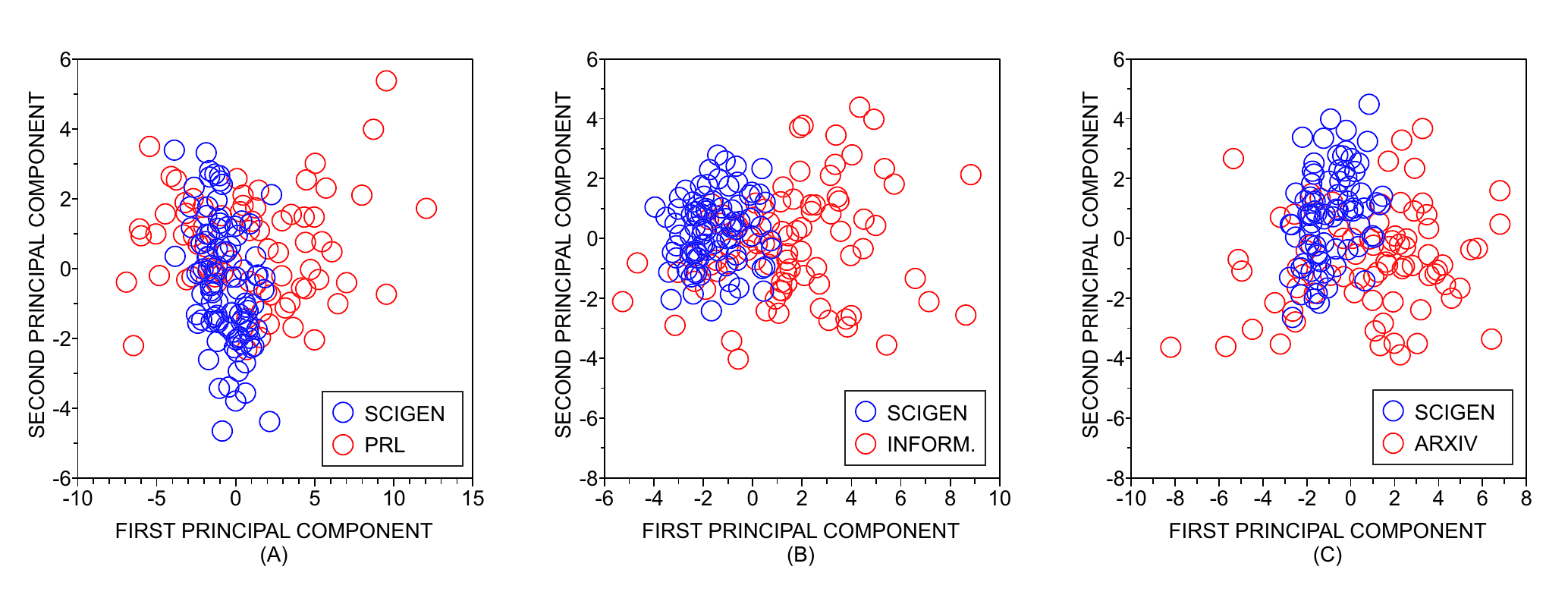}
\caption{Principal component analysis obtained with all measurements. The highest discriminability, as revealed by the size of the overlapping regions, occurs for the J. Inform. and arXiv datasets. The topological variability of the automatically generated texts from SCIGen tends is lower than the variability observed in real manuscripts.}
\label{afigpca}
\end{figure}

In the multivariate analysis, all 13 measurements were combined and applied as features of the classifiers. A two-dimensional projection of the data using the principal component analysis technique~\cite{livrodefrente} is shown in Figure \ref{afigpca}. Interestingly, it is possible to note that the worst discrimination occurred in the PRL dataset, as revealed by the large overlapping region. Conversely, a much better discrimination was achieved with the arXiv.org/cs dataset. These results are consistent with the patterns found when the univariate analysis was performed.
Another interesting pattern arising from the visualization provided in Figure \ref{afigpca} concerns the variability of style of SCIGen papers. It is clear that the style of SCIGen papers displays a lower variability when compared to the style of real texts. This effect can be easily perceived, e.g. by observing that SCIGen papers are scattered in a small region in Figure \ref{afigpca}(b).

The accuracy rates obtained with the multivariate classification is shown in Table \ref{atab1}. When one compares the results obtained here with the ones achieved with  the univariate analyses, it is clear that the multivariate analysis improved the discriminative ability of the classifiers. The accuracy rate in the PRL dataset improved 10\% (from 79\% to 89\%). In the arXiv/cs dataset, the accuracy went from 88\% to 95\%. The lowest increase in accuracy occurred for the JI dataset, which already had provided an excellent discriminability with the univariate approach. These results suggest that the interaction of attributes is able to improve the identification of fake papers generated by the SCIGen algorith, especially if the separation between real and fake papers is not so clear when a single measurement is employed to generate the classifiers.
\begin{table}
\centering
    \caption{\label{atab1}Accuracy rate (\%) obtained when distinguishing real (PRL, arXiv/cs or JI) from artificial papers (SCIGen). Unlike the univariate analysis, all 13 topological measurements were employed.}
    \begin{tabular}{|l|c|c|c|}
      \hline
      {\bf Dataset} & {\bf KNN} & {\bf NBY} & {\bf C45} \\
      \hline
      PRL   & 83 & 89 & 85 \\
      arXiv.org/cs       & 95 & 94 & 87 \\
      JI      & 95 & 95 & 95 \\
      \hline
    \end{tabular}
\end{table}

\begin{figure}[!h]
\centering
\includegraphics[width=0.65\linewidth]{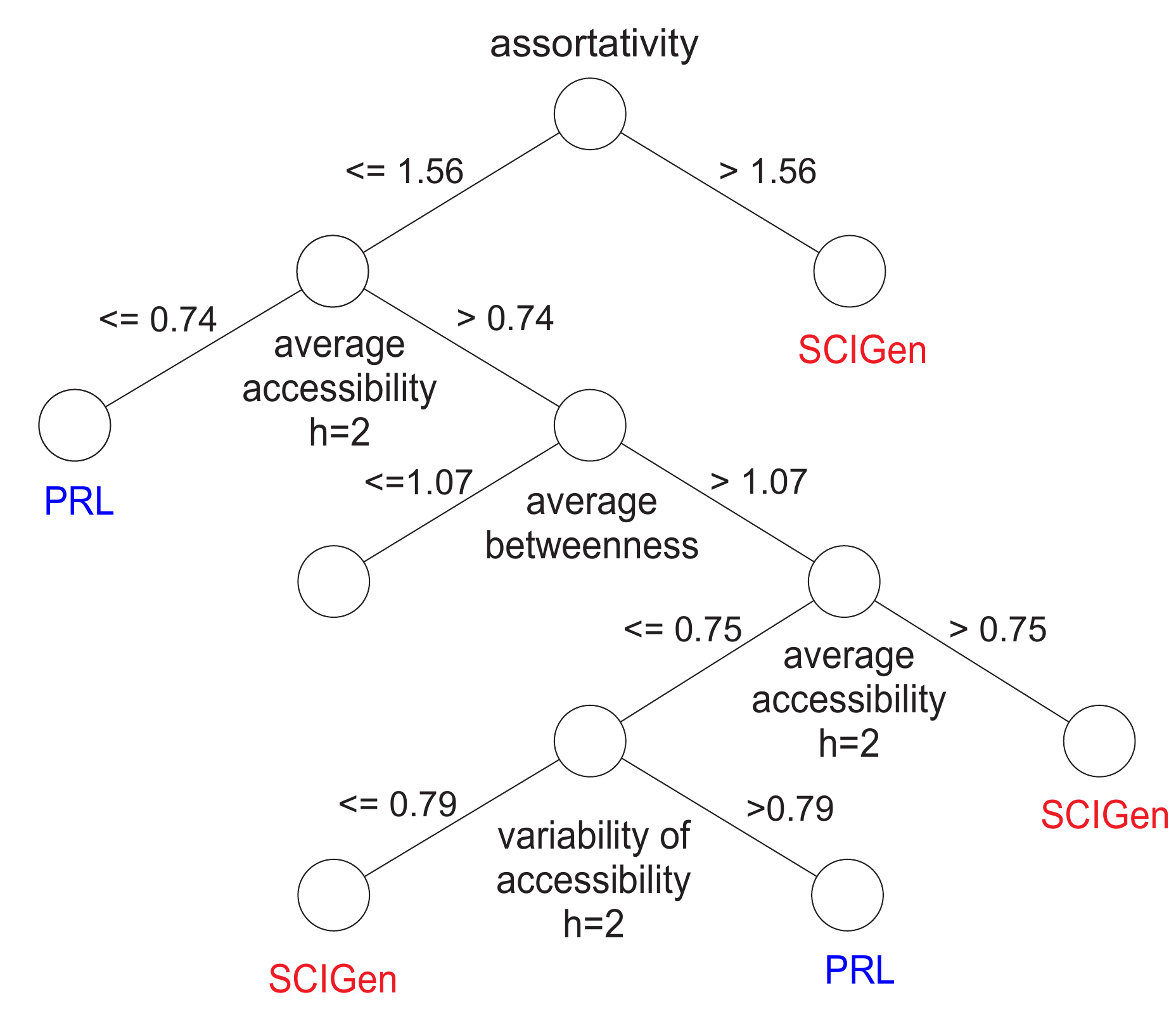}
\caption{Decision tree obtained for distinguish PRL from SCIGen manuscripts. This decision tree is able to accurately identify the class (PRL or SCIGen) of the manuscripts in 85\% of the cases. Note that not all measurements were employed for the classification.}
\label{fig.aarvored}
\end{figure}

Although all attributes have been used as input to the machine learning methods, only some of them are selected to generate a given model. This is clear when one observes the decision tree shown in Figure \ref{fig.aarvored}, which summarizes the patterns recognized in the PRL dataset. %
The relevance of each attribute employed in the multivariate analysis was quantified with the technique described in Section \ref{qfr}. The ranking obtained for each measurement in each classifier is shown in Table \ref{segtab}. Note that there is a strong consistency between rankings of measurements across distinct classifiers in the same dataset. This consistency is confirmed by the high values of {Spearman's rank correlation of rankings (see Table \ref{spearmantable})}. The performance of each measurement for the classification is commented below.

\begin{table}
    \centering
    \caption{\label{spearmantable}Spearman's rank correlation coefficient for distinct rankings of attributes. Note that, in general, there is a strong correlation between the rankings obtained in the same dataset.}
    \begin{tabular}{|l|c|c|c|}
      \hline
      {\bf Dataset} & {\bf PRL} & {\bf arXiv/cs} & {\bf JI}\\
      \hline
      KNN and NBY & 0.703 & 0.703 & 0.709 \\
      KNN and C45 & 0.648 & 0.698 & 0.916 \\
      NBY and C45 & 0.192 & 0.654 & 0.640 \\
      \hline
    \end{tabular}
\end{table}

\begin{table}
    \centering
    \caption{\label{segtab}Ranking of measurements based on the accuracy rates of the classifiers, where 1 means best, 2 second best and so forth. In this analysis, the multiple interactions between features was considered.  The results obtained for each classifier is showed for each dataset considered. Note that, in general, the performance depends on the dataset.}
    \begin{tabular}{|l|c|c|c|c|c|c|c|c|c|}
      \hline
       & \multicolumn{3}{|c|}{\bf P. Rec. Lett.} & \multicolumn{3}{|c|}{\bf arXiv.org/cs} & \multicolumn{3}{|c|}{\bf J. Informetr.} \\
       \cline{2-10}
       & KNN & NBY & C45 & KNN & NBY & C45 & KNN & NBY & C45 \\
      \hline
      Accessibility $\langle \alpha^{(h=2)} \rangle$ & 3  & 4  & 5  & 7  & 10 & 10 & 3  & 6  & 3 \\
      Accessibility $\Delta \alpha^{(h=2)}$          & 9  & 9  & 8  & 12 & 12 & 5  & 13 & 11 & 15 \\
      Accessibility $\langle \alpha^{(h=3)} \rangle$ & 4  & 5  & 3  & 4  & 8  & 9  & 4  & 7  & 6 \\
      Accessibility $\Delta \alpha^{(h=3)}$          & 10 & 10 & 9  & 1  & 1  & 1  & 1  & 1  & 1 \\
      Avg. N. Degree $\langle k_n \rangle$          & 5  & 6  & 2  & 5  & 9  & 6  & 5  & 8  & 5 \\
      Avg. N. Degree $\Delta k_n$                  & 11 & 11 & 10 & 2  & 2  & 2  & 2  & 2  & 2 \\
      Betweenness $\langle B \rangle$               & 2  & 1  & 1  & 3  & 3  & 3  & 6  & 5  & 7 \\
      Betweenness $\Delta B$                        & 1  & 3  & 6  & 6  & 4  & 4  & 9  & 3  & 12 \\
      Clustering $\langle C \rangle$                & 8  & 8  & 11 & 9  & 6  & 7  & 10 & 10 & 8 \\
      Clustering $\Delta C $                        & 12 & 7  & 12 & 10 & 7  & 8  & 11 & 9  & 10 \\
      Assortativity $r$                             & 7  & 2  & 13 & 11 & 5  & 13 & 7  & 4  & 4 \\
      Shortest paths $\langle l \rangle$            & 6  & 13 &  4 & 8  & 11 & 11 & 8  & 13 & 9 \\
      Shortest paths $\Delta l$                     & 13 & 12 &  7 & 13 & 13 & 12 & 12 & 12 & 11 \\
      \hline
    \end{tabular}
\end{table}

\begin{itemize}

  \item {\bf Accessibility}: the performance of this measurement depends on the dataset. The deviation $\Delta \alpha^{(h=3)}$ turned out to be the best measurement to identify real papers in the arXiv/cs and JI datasets. Differently, in the PRL dataset, the average   $\langle \alpha^{(h=3)} \rangle$ performed better than the deviation $\Delta \alpha^{(h=3)}$. The best performance using accessibility measurements in the PRL dataset was achieved with the average $\langle \alpha^{(h=2)} \rangle$.

  \item {\bf Neighbors degree}: an excellent performance was observed for the deviation $\Delta k_n$ in both arXiv/cs and JI datasets. Note that $\Delta k_n$  reached second place in both repositories. Particularly, in the PRL dataset, the average $\langle k_n \rangle$ performed better than the deviation $\Delta k_n$.

  \item {\bf Betweenness}: the average $\langle B \rangle$ performed very well in all three datasets. This suggests that this measurement becomes very discriminative when combined with other attributes. Note that, in the univariate analysis, the betweenness displayed low accuracy rates (see Table \ref{univ.result}).

  \item {\bf Clustering coefficient}: the combination with other attributes does not seem to improve the discriminability of this measurement.

  \item {\bf Assortativity}: the importance of this measurement depends on the dataset. The best performance, a second position, was achieved in the PRL dataset when the Naive Bayes classifier was used.

  \item {\bf Shortest paths}: the average $\langle l \rangle$  and specially the deviation $\Delta l$ ranked among the worst measurements. Therefore, similarly to the clustering coefficient, the average shortest path length is not informative even when associated with others measurements.

\end{itemize}

All in all, the combination of attributes improved the performance of the classifications.
The attributes with the highest discrimination ability were the average betweenness $\langle B \rangle$ (PRL dataset) and the standard deviation of the accessibility $\Delta \alpha^{(h=3)}$.
Although some measurements turned out to be not informative in specific datasets, they still can be useful in other scenarios, as the discriminability may depend on the data distribution.
For this reason, the clustering coefficient and the average shortest path length should be tried in other datasets.

\section{Conclusions} \label{sconclusion}

{In the current paper, I have investigated the hypothesis that artificially generated manuscripts can be distinguished from real scientific papers via topological characterization of complex networks. The combination of network features (extracted from the word adjacency model) and machine learning methods allowed the correct identification of SCIGen papers in 89\% of the cases (worst scenario). \Red{This means that there are hidden patterns in the organization of papers generated by SCIGen that differs from the structural patterns arising from real texts. Even though the techniques presented in this manuscript does not outperform the methods based on textual content, it could be employed in applications where the complementary nature of the proposed attributes plays a prominent role to discriminate pieces of texts with similar content~\cite{lang6,dis1}}.

The analysis of relevance of attributes revealed that the combination of distinct topological attributes is the most successful approach. Concerning the individual performance of topological features, the accessibility and the betweenness performed particularly well mainly in the multivariate analysis. Conversely, the clustering coefficient and the shortest path length displayed the poorest performance among the topological features employed. The results presented here confirm, as a proof of principle, that the word adjacency model can be useful to identify fake papers. Future works could pursue an improvement of performance with a fine tuning of classifiers parameters~\cite{sistematica}. \Red{Another possibility is to propose novel topological measurements to combine the techniques presented in this paper with traditional statistical natural language processing methods~\cite{statnlp}.}

\begin{acknowledgements}
I am thankful to S\~ao Paulo Research Foundation (FAPESP) (grant number 14/20830-0) for the financial support.
\end{acknowledgements}



\begin{thebibliography}{00}

\bibitem{evaluacao}
Radicchi, F., Fortunato, S., Markines, B., \& Vespignani, A. (2009)
Diffusion of scientific credits and the ranking of scientists.
Phys. Rev. E 80, 056103.



\bibitem{acorrelacao}
Finardia, U. (2013)
Correlation between journal impact factor and citation performance: an experimental study.
Journal of Informetrics, 7(2) 357--370.

\bibitem{ex1}
Glanzel, W., Schlemmer, B., \& Thijs, B. (2003) Better late than never? On the chance to become highly cited only beyond the standard time horizon. Scientometrics 58(3), 571--586.

\bibitem{ex2}
Peirce, C.S. (1884) The numerical measure of the success of predictions.
Science 4(93), 453--454.

\bibitem{ex3}
Van Calster, B. (2012) It takes time: a remarkable example of delayed recognition.
Journal of the American Society for Information Science and Technology 63(11), 2341--2344.

\bibitem{ex4}
Wu, Y., Fu, T.Z.J., \& Chiu, D.M. (2014)
Generalized preferential attachment considering aging.
Journal of Informetrics 8 (3), 650--658.

\bibitem{ex5}
Hajra, K.B., \& Sen P. (2005)
Aging in citation networks.
Physica A 346 (1--2), 44--48.

\bibitem{selfie1}
Ferrara, E., \& Romero, A. E. (2013)
Scientific impact evaluation and the effect of self-citations: mitigating the bias by discounting the h-index.
Journal of the American Society for Information Science and Technology 64(11), 2332--2339.

\bibitem{selfie2}
Yua, T., Yua, G., \& Wang M-Y. (2014) Classification method for detecting coercive self-citation in journals.
Journal of Informetrics 8(1), 123--135.

\bibitem{plagio1}
Glänzel, W., Braun, T., \& Schubert, A., Zosimo-Landolfo, G. (2014)
Scientometrics 102(1), 1--3.

\bibitem{plagio2}
García-Romero, A., \& Estrada-Lorenzo, J. M. (2014)
A bibliometric analysis of plagiarism and self-plagiarism through Déjà vu.
Scientometrics 101(1), 381--396.

\bibitem{plagio3}
Citron, D. T., \& Ginsparg, P. (2015)
Patterns of text reuse in a scientific corpus.
PNAS 112(1), 25--30.

\bibitem{labbe2010}
\Red{
Labb\'e, C. (2010).
Ike antkare, one of the great stars in the scientific firmament.
International Society for Scientometrics and Informetrics Newsletter 6(2), 48--52.
}

\bibitem{sciref}
\url{pdos.csail.mit.edu/scigen}

\bibitem{hindman}
Bartneck, C., \& Kokkelmans, S. (2011)
Detecting h-index manipulation through self-citation analysis.
Scientometrics 87(1), 85--98.

\bibitem{newman}
Newman, M. (2010)
Networks: An Introduction. Oxford University Press, Inc., New York, NY, USA.


\bibitem{xiang}
\Red{
Xiong, J., \& Huang, T. (2009)
An effective method to identify machine automatically generated paper.
In Pacific-Asia Conference on Knowledge Engineering and Software Engineering, 101--102.
}


\bibitem{ginsparg}
\Red{
Ginsparg, P. (2014)
Automated screening: arXiv screens spot fake papers.
Nature 508(7494): 44.
}


\bibitem{noorden}
\Red{
Van Noorden, R. (2014)
Publishers withdraw more than 120 gibberish papers.
Nature 24.
}

\bibitem{lavoie}
\Red{
Lavoie, A., \& Krishnamoorthy, M. (2010).
Algorithmic detection of computer generated text.
arXiv: abs/1008.0706
}

\bibitem{statnlp}
Manning, C.D., \& Schutze, H. (1999)
Foundations of Statistical Natural Language Processing. MIT Press, Cambridge, MA, USA.


\bibitem{labbe2013}
\Red{
Labb\'e, C., \& Labb\'e, D. (2013).
Duplicate and fake publications in the scientific literature: how many scigen papers in computer science?
Scientometrics 94(1): 379--396.}

\bibitem{estrutural}
\Red{
Fahrenberg, U., Biondi, F., Corre, K., Jégourel, C., Kongshoj, S., \& Legay, A. (2014)
Measuring structural distances between texts.
arXiv: abs/1403.4024.
}

\bibitem{li}
\Red{
Li, M., Chen, X., Li, X., Ma, B., \& Vitanyi, P. (2004)
The similarity metric.
IEEE Transactions on Information Theory 50(12), 3250--3264.
}

\bibitem{compressao}
\Red{
Dalkilic, M. M., Clark, W. T.,Costello, J. C., \& Radivojac, P. (2006)
Using compression to identify classes of inauthentic texts.
In Proceedings of the 2006 SIAM Conference on Data Mining.
}

\bibitem{extractive}
Amancio, D.R., Nunes, M.G.V., Oliveira Jr., O.N., \& Costa, L. da F. (2012)
Extractive summarization using complex networks and syntactic dependency.
Physica A, 391 1855--1864.
%
\bibitem{lantiq}
Antiqueira, L., Oliveira Jr., O. N., Costa, L. da F., \& Nunes, M. G. V. (2009) A complex network approach to text summarization.
Information Sciences, 179, 584--599.

\bibitem{trad1}
Amancio, D.R., Antiqueira, L., Pardo, T.A.S., Costa, L. da F., Oliveira Jr., O.N., \& Nunes, M.G. V. (2008)
Complex networks analysis of manual and machine translations.
International Journal of Modern Physics C 19, 583--598.

\bibitem{trad2}
Amancio, D.R., Nunes, M.G.V., Oliveira Jr., O.N., Pardo, T.A.S., Antiqueira, L., \& Costa, L. da F. (2011)
Using metrics from complex networks to evaluate machine translation. Physica A 390, 131--142.

\bibitem{complexity1}
Amancio, D.R., Aluisio, S.M., Oliveira Jr., O.N.,  \& Costa, L. da F. (2012) Complex networks analysis of language complexity.
EPL 100, 58002.

\bibitem{complexity2}
Yasseri, T., Kornai, A., \& Kert\'esz, J. (2012)
A practical approach to language complexity: a wikipedia case study.
PLoS ONE 7, e48386.

\bibitem{complexity3}
Liu, H., \& Xu, C. (2011) Can syntactic networks indicate morphological complexity of a language?
EPL 93, 28005.

\bibitem{complexity4}
Sol\'e, R.V., Corominas-Murtra, B.B., Valverde, S. \& Steels, L. (2009)
Language networks: their structure, function and evolution.
Complexity 15(6), 20--26.

\bibitem{complexity5}
Liu, H. (2008)
The complexity of chinese syntactic dependency networks.
Physica A 387, 3048--3058.

\bibitem{njp}
Amancio, D.R., Altmann, E.G.,  Oliveira Jr., O.N., \& Costa, L. da F. (2011)
Comparing intermittency and network measurements of words and their dependency on authorship.
New J. Phys. 13, 123024.

\bibitem{plosrecent}
Amancio, D.R. (2015)
Probing the topological properties of complex networks modeling short written texts.
PLoS ONE 10 e0118394. DOI: 10.1371/journal.pone.0118394.

\bibitem{probing}
Amancio, D.R., Altmann, E.G., Rybski, D., Oliveira Jr., O. N., \& Costa, L. da F. Probing the statistical properties of unknown texts: application to the Voynich manuscript.
PLOS ONE 8, p. e67310, 2013.

\bibitem{lang2}
Mota, N.B., Furtado, R., Maia, P.P.C, Copelli, M., \& Ribeiro, S. (2014) Graph analysis of dream reports is especially informative about psychosis. Scientific reports 4, 3691.



\bibitem{lang3}
Ronen, S., Gonçalves, B., Hu, K.Z., Vespignani, A., Pinker, S., \& Hidalgo, C.A. (2014)
Links that speak: the global language network and its association with global fame. PNAS 111(52), 5616--5622.

\bibitem{lang4}
Baronchelli, A., Ferrer-i-Cancho, R., Pastor-Satorras, R., Chater, N., \& Christiansen, M.H. (2013)
Networks in cognitive science.
Trends in cognitive sciences 17, 348-60.

\bibitem{lang5}
Masucci, A.P., Kalampokis, A., Eguíluz, V.M., \& Hernández-García, E. (2011)
Wikipedia information flow analysis reveals the scale-free architecture of the semantic space.
PLoS ONE 6(2), e17333.

\bibitem{lang6}
Silva, T.C., \& Amancio, D.R. (2013)
Discriminating word senses with tourist walks in complex networks.
The European Physical Journal B 86, 297.


\bibitem{doliu1}
\Red{
Liu, H., \& Li, W. (2010)
Language clusters based on linguistic complex networks.
Chinese Sci. Bull. 55(30): 3458--3465.
}

\bibitem{domehler}
\Red{
Abramov, O., \& Mehler, A. (2011)
Automatic language classification by means of syntactic dependency networks.
J. Quant. Linguist. 18(4): 291--336.
}

\bibitem{doliu2}
\Red{
Liu, H.T. \& Cong, J. (2013)
Language clustering with word co-occurrence networks based on parallel texts.
Chinese Sci. Bull. 58(10): 1139--1144.
}

\bibitem{lang1}
Cong J, \& Liu, H. (2014)
Approaching human language with complex networks.
Physics of Life Reviews 11(4), 598--618.



\bibitem{sigman}
Sigman, M., \& Cecchi, G.A. (2002)
Global organization of the Wordnet lexicon.
PNAS 99(3), 1742--1747.

\bibitem{syntactic}
Ferrer i Cancho, R., Solé, R.V., \& Kohler, R. (2004)
Patterns in syntactic dependency networks.
Physical Review E 69, 051915.

\bibitem{lemmatisation}
Liu, H., Christiansen, T., Baumgartner, W.A., \& Verspoor, K. (2012) BioLemmatizer: a lemmatization tool for morphological processing of biomedical text. Journal of Biomedical Semantics 3: 3.


\bibitem{maxent}
Berger, A.L., Della Pietra, V.J., \& Della Pietra, S.A. (1996)
A maximum entropy approach to natural language processing.
Comput. Linguist., 22(1), 39--71.

\bibitem{border}
Traven\c{c}olo, B.A.N., \& Costa, L. da F. (2008)
Accessibility in complex networks.
Phys. Lett. A 373, 89--95.

\bibitem{identification}
Amancio, D.R., Oliveira Jr., O.N., \& Costa, L. da F. (2012) Identification of literary movements using complex networks to represent texts.
New J. Phys. 14, 043029.









\bibitem{livrodefrente}
Costa, L. da F. (2014) Shape classification and analysis: theory and practice. CRC Press, 2 edition.

\bibitem{mixing}
Newman, M.E.J. (2003)
Mixing patterns in networks.
Phys. Rev. E 67, 026126.

\bibitem{finding}
Newman, M.E.J. (2006)
Finding community structure in networks using the eigenvectors of matrices.
Physical Review E 74, 036104.

\bibitem{parzen}
Parzen, E. (1962)
On estimation of a probability density function and mode.
The Annals of Mathematical Statistics 33(3), 1065.

\bibitem{sistematica}
Amancio, D.R., Comin, C.H., Casanova, D., Travieso, G., Bruno, O.M., Rodrigues, F.A., \& Costa, L. da F. (2014) A systematic comparison of supervised classifiers. PLOS ONE 9, e94137.

\bibitem{cormen}
Cormen, T.H., Stein, C., Rivest, R.L., \&  Leiserson, C.E. (2001) Introduction to Algorithms, McGraw-Hill Higher Education.

\bibitem{duda}
Duda, R.O., Hart, P.E., \& Stork, D.G. (2000)
Pattern Classification (2nd Edition). Wiley-Interscience.

\bibitem{prlj}
\url{journals.elsevier.com/pattern-recognition-letters}

\bibitem{arxivj}
\url{www.arXiv.org/archive/cs}

\bibitem{jinfor}
\url{journals.elsevier.com/journal-of-informetrics}


\bibitem{dis1}
Silva, T.C., \& Amancio, D.R. (2012)
Word sense disambiguation via high order of learning in complex networks.
EPL 98, 58001.

\end{thebibliography}

\end{document}